\pdfoutput=1

\documentclass{sig-alternate-05-2015}
\usepackage[caption=false]{subfig}
\usepackage{amsmath}
\usepackage{xcolor}
\usepackage[ruled,vlined]{algorithm2e}
\usepackage{xfrac}
\usepackage{cite}
\usepackage[normalem]{ulem}
\newcommand{\specialcell}[2][c]{%
  \begin{tabular}[#1]{@{}c@{}}#2\end{tabular}}

\begin{document}






\title{Universum Learning for SVM Regression}

%
%
%
%
%

\numberofauthors{2} 
%
\author{
%
%
\alignauthor
Sauptik Dhar\titlenote{Majority of this work was done during S.Dhar's PhD, and is available in his thesis: \url{http://conservancy.umn.edu/handle/11299/162636}}\\
       \affaddr{Research and Technology Center}\\
       \affaddr{Robert Bosch LLC}\\
       \affaddr{Palo Alto, CA 94304, USA}\\
       \email{sauptik.dhar@us.bosch.com}
\alignauthor
Vladimir Cherkassky \\
       \affaddr{Department of ECE}\\
       \affaddr{University of Minnesota}\\
       \affaddr{Minneapolis, MN 55455, USA}\\
       \email{cherk001@umn.edu}
}

\maketitle
\begin{abstract}

This paper extends the idea of Universum learning \cite{vapnik98,vapnik06} to regression problems. We propose new Universum-SVM formulation for regression problems that incorporates a priori knowledge in the form of additional data samples. These additional data samples or Universum belong to the same application domain as the training samples, but they follow a different distribution. Several empirical comparisons are presented to illustrate the utility of the proposed approach.

\end{abstract}

%
%


%
%

%
%



\section{Introduction} \label{sec1}
The technique of Universum learning or \textit{learning through contradiction} \cite{vapnik98,vapnik06} provides a formal mechanism for incorporating a priori knowledge about the application domain, in the form of additional (unlabeled) Universum samples. Universum learning has been originally introduced for binary classification problems \cite{weston06,vapnik06} and it has been shown to be particularly effective for high-dimensional low-sample size data settings \cite{weston06,cherkassky11,sinz08}. More recently, Universum learning has been extended to various non-standard classification settings \cite{chen09,dhar15,lu14,qi14,shen12,wang14,zhang08}. However, most research on Universum learning has been limited to binary classification problems. It is not clear how to extend or modify the Universum learning approach to other types of learning problems.

Besides classification setting, another common supervised learning problem is \textit{regression} or real-valued function estimation from noisy samples \cite{cherkassky13,cherkassky07}. The output in regression problems is a random variable that takes on real values and can be interpreted as the sum of a deterministic function $t(\mathbf{x})$ and a random error $\delta$ with zero mean:
\begin{equation}\label{eq1}
	y = t(\mathbf{x}) + \delta 
\end{equation}
where, the deterministic part, aka the target function, is the mean of the output conditional probability,
\begin{equation}\label{eq2}
t(\mathbf{x}) = \int yp(y\mid \mathbf{x})\ dx
\end{equation}
Here, $(\mathbf{x},y)$ follows an underlying distribution described by the joint density function,
\begin{equation}\label{eq3} 
 p(\mathbf{x}, y)= p(y\mid \mathbf{x}) p(\mathbf{x})	    
\end{equation}
The goal of the learning problem is to estimate the `\textit{best}' function (model) from a set of approximating functions  $f(\mathbf{x},\omega)$ parameterized by $\omega \in \Omega $. The quality of the approximation is measured by a loss or discrepancy function $L(y,f(\mathbf{x},\omega)) $. Thus, the problem of regression involves estimation of a real-valued function that minimizes the risk functional,
\begin{equation}\label{eq4}
R(\omega)= \int L(y,f(\mathbf{x},\omega)) p(\mathbf{x}, y) d\mathbf{x}dy
\end{equation}
A typically chosen loss function for most regression problems is the squared loss, 
\begin{equation}\label{eq5}
L(y,f(\mathbf{x},\omega)) = (y-f(\mathbf{x},\omega))^2 
\end{equation}           	          
The difficulty of regression estimation is due to the fact that statistical distribution of $(\mathbf{x},y)$ is unknown, and the only information (about it) is available in the form of finite training data set $(\mathbf{x}_i,y_i)_{i=1}^n$. For regression problems, one can also expect to achieve improved generalization performance by incorporating a priori knowledge in the form of additional Universum samples. However, formalizing the notion of `contradiction' for regression setting is not straightforward. This paper describes the concept of Universum learning for regression problems and provides new optimization formulation that incorporates additional Universum data into SVM regression setting.

This paper is organized as follows. Section \ref{sec2} describes standard SVM regression (SVR) formulation. Section \ref{sec3} extends the notion of Universum learning for regression problems and introduces the new Universum-SVM regression (U-SVR) formulation. Next we provide empirical results to illustrate the effectiveness of this new formulation in Section \ref{sec4}. Finally, Section \ref{sec5} presents conclusions.

\section{SVM Regression} \label{sec2}
\begin{figure}
\centering
\subfloat[][]{\includegraphics[height=1.4in, width=1.5in]{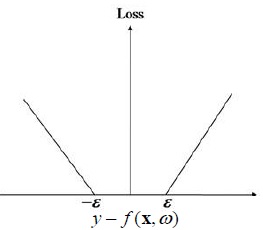}\label{fig1a}}
\subfloat[][]{\includegraphics[height=1.4in, width=1.5in]{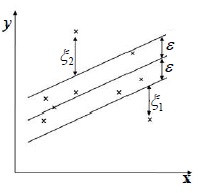}\label{fig1b}}
    \caption{SVM regression (a) $\varepsilon$ - insensitive loss function. (b) slack variables $\xi$ for linear SVM regression formulation.}
\end{figure}

This section provides brief description of standard SVM regression (SVR) formulation following Vapnik \cite{vapnik98,vapnik06}. This SVM regression setting employs special margin-based loss function known as $\varepsilon$ -insensitive loss $L_{\varepsilon}(f(\mathbf{x},\mathbf{w}),y)=$ $max(\mid y-f(\mathbf{x},\mathbf{w})\mid-\varepsilon,0) $. Note that this loss function is defined in the space of the target values $y \in \Re$, as shown in  Fig. 1. Then SVM regression optimization formulation (for linear parameterization) can be stated as follows: 

Given i.i.d training samples $(\mathbf{x}_{i},y_{i})_{i=1}^{n}$; with  $\mathbf{x} \in \Re^{d}$, $y \in \Re$, the linear SVM model can be found by solving the optimization problem

\begin{flalign}\label{eq6}
\underset{\mathbf{w},b}{\text{min}}&\quad \quad  \frac{1}{2}(\mathbf{w}\cdot \mathbf{w}) + C\sum\limits_{i=1}^n(\xi_{i} +\xi_{i}^{*}) &&\\ 
s.t. &\quad \quad  y_{i}-(\mathbf{w}\cdot \mathbf{x_{i}})-b \leq \varepsilon+\xi_{i} \quad  \xi_{i}\geq 0 \quad i=1 \ldots n \nonumber\\ 
& \quad \quad (\mathbf{w}\cdot \mathbf{x}_{i})+b-y_{i} \leq \varepsilon+\xi_{i}^{*} \quad  \xi_{i}^{*}\geq 0 \quad i=1 \ldots n \nonumber
\end{flalign}
here $n$ := number of training samples, and $d$ := dimensionality of the input space (or the number of input variables). Note that training samples falling inside the $\varepsilon$ - tube have zero loss, and samples outside the $\varepsilon$ - insensitive zone are linearly penalized using the slack variables $\xi_{i},\xi_{i}^{*} \geq 0$, $i=1 \ldots n$ (as shown in Fig. \ref{fig1b}). These slack variables contribute to the empirical risk for the SVR formulation $R_{emp}(\mathbf{w})=\sum \limits_{i=1}^{n}(\xi_{i} + \xi_{i}^{*}) $. The SVR formulation attempts to strike a balance between the minimization of empirical risk and the penalization term. The user-defined parameter $C \geq 0$ controls the trade-off between the empirical risk and the penalization term, and $\varepsilon \geq 0 $  controls the size of the $ \varepsilon $ - insensitive zone in margin-based loss. Both of these parameters jointly control the SVM model complexity and hence its generalization performance.  For most SVR solvers, problem \eqref{eq6} is usually solved in its dual form (see \cite{cherkassky13} for details):

\begin{flalign}\label{eq7}
\underset{\mathbf{\alpha},\mathbf{\beta}}{\text{min}} & \quad \quad \varepsilon \sum \limits_{i=1}^{n}(\alpha_{i} + \beta_{i}) - \sum \limits_{i=1}^{n} y_{i}
(\alpha_{i} - \beta_{i})&&\\
& \quad \quad + \frac{1}{2}\sum \limits_{i,j =1}^n (\alpha_{i} - \beta_{i})(\alpha_{j} - \beta_{j})(\mathbf{x}_{i} \cdot \mathbf{x}_{j})  \nonumber \\
s.t. & \quad \quad \sum \limits_{i=1}^{n} \alpha_{i} = \sum \limits_{i=1}^{n} \beta_{i}, \quad 0 \leq \alpha_{i},\beta_{i} \leq C, \quad i = 1 \ldots n    \nonumber
\end{flalign}

Solution of the dual formulation \eqref{eq7} yields optimal values of parameters $(\alpha_{i}^{*},\beta_{i}^{*})_{i=1}^{n}$ that can be used to construct the optimal SVR function: $f(\mathbf{x})= \sum \limits_{i=1}^n (\alpha_{i}^{*} - \beta_{i}^{*})(\mathbf{x}_{i} \cdot \mathbf{x})+b$. In this optimal solution, training samples with non-zero coefficients are the support vectors (SVs), corresponding to data points at the boundary or outside $\varepsilon$-insensitive zone.

This dual formulation \eqref{eq7} can be also used to extend linear SVR to a non-linear setting. This is accomplished by replacing the dot product $(\mathbf{x}_i \cdot \mathbf{x}_j)$ in \eqref{eq7} with a non-linear kernel $K(\mathbf{x}_i,\mathbf{x}_j)$. This kernel $K(\mathbf{x}_i,\mathbf{x}_j) =( \varphi(\mathbf{x}_i) \cdot \varphi(\mathbf{x}_j))$  implicitly captures the non-linear mapping of the data $\mathbf{x} \rightarrow \varphi(\mathbf{x})$. Commonly used kernel functions include:
\begin{itemize}
\item[--] Polynomial kernel (of degree $q$): 
\item[]\quad \quad $K(\mathbf{x}_i,\mathbf{x}_j) = ((\mathbf{x}_i \cdot \mathbf{x}_j)+1)^q$
\item[--] Radial Basis Function (RBF) (with parameter $\gamma$): 
\item[]\quad \quad $K(\mathbf{x}_i,\mathbf{x}_j) = exp(-\gamma \Vert \mathbf{x}_i - \mathbf{x}_j \Vert^2)$
\end{itemize} 
For more details see \cite{cherkassky13,drucker97}.

\section{Universum-SVM Regression} \label{sec3}

\begin{figure} 
\centering
\includegraphics[height=1.5in, width=2in]{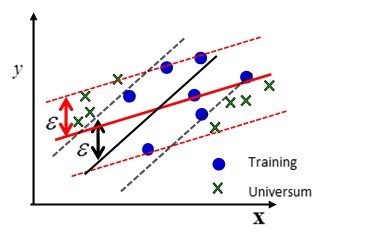} 
\caption{Two SVM regression models explain training data equally well, but have different number of contradictions on the Universum. The model with a larger number of contradictions (in black) is selected.} \label{fig2}
\end{figure} 

\subsection{Universum-SVM Regression formulation} \label{sec31}

\begin{figure}
\centering
\includegraphics[height=1.6in, width=2.1in]{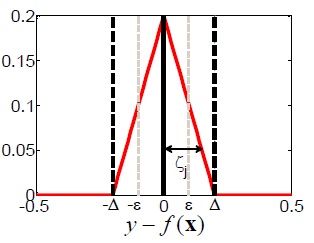}
\caption{Loss function for the universum samples $U_{\Delta}(y_{j}^{*}-f(\mathbf{x}_{j}^{*}))$ (with $\Delta$ = 0.2 for illustration).} \label{fig3}
\end{figure}

This section describes proposed Universum-SVM regression formulation, using \textit{new notion of falsification} for regression setting, as explained next. Consider regression setting where available training data $(\mathbf{x}_i,y_i)_{i=1}^n $ is modeled using linear SVR. As described in Section \ref{sec2}, for SVM regression the concept of `margin' is implemented via $\varepsilon$ - insensitive zone (Fig. \ref{fig1a}). That is, training samples falling inside $\varepsilon$ - insensitive zone are `explained' by SVM model, and samples falling outside `falsify' or `contradict' this model. Next, consider two SVR models which explain training samples \textit{equally well}, e.g. both SVR models use the same value of $\varepsilon$ and achieve the same empirical risk  $R_{emp}(\mathbf{w})=\sum \limits_{i=1}^{n}(\xi_{i} + \xi_{i}^{*}) $ for training samples. For example, Fig. \ref{fig2} shows two SVR models that explain available training data equally well, i.e. have zero empirical risk. Now, consider additional \textit{Universum} samples $(\mathbf{x}_j^{*},y_j^{*})_{j=1}^m$. These samples are defined in the same $(\textbf{x}, y)$ space as the training samples, and they reflect a priori knowledge that they should not be explained well by SVM regression model. That is, Universum samples should \textit{lie outside} the $\varepsilon$ - insensitive tube. For the toy example shown in Fig. \ref{fig2}, we should favor the model shown in black, for which most Universum samples cannot be explained by SVR model. Note that for regression setting, the Universum samples are \textit{labeled}, unlike unlabeled Universum samples for classification. This reasoning motivates new Universum support vector regression (U-SVR) formulation where:

\begin{itemize}
\item[--] Standard  $\varepsilon$ - insensitive loss is used for training samples. This loss forces training samples to lie inside  $\varepsilon$ - insensitive tube. 
\item[--] Universum samples are penalized by a different loss function as shown in Fig. \ref{fig3}.
\end{itemize} 
This new loss function forces the universum samples to lie `far away' from the regression model, so that samples outside a  $\pm \Delta $ zone have zero loss. Penalization of universum samples inside the $\pm \Delta $ zone is achieved via the slack variables $\zeta_j$ as shown in Fig. \ref{fig3}. Note that the tunable parameter $ \Delta $ can be larger (or smaller) than $\varepsilon$. This leads to the following optimization formulation for U-SVR:

\begin{flalign} \label{eq8}
\underset{\mathbf{w},b,\mathbf{\xi},\mathbf{\xi}^{*},\mathbf{\zeta}}{\text{min}}&\quad L(\mathbf{w},b,\mathbf{\xi}_{i},\mathbf{\xi}^{*},\mathbf{\zeta}) = &&
\end{flalign}
\quad\textcolor{blue}{\textbf{(Training samples)}}\quad \quad \quad \quad \textbf{(Universum samples)}
\begin{equation}
\begin{split}
\boxed{
\begin{array}{lll}
\frac{1}{2}(\mathbf{w}\cdot \mathbf{w}) + C\sum\limits_{i=1}^n(\xi_{i} +\xi_{i}^{*})\hspace{-0.8em}\\ 
s.t. \\
y_{i}-(\mathbf{w}\cdot \mathbf{x}_{i})-b \leq \varepsilon+\xi_{i} \hspace{-0.8em}\\ 
(\mathbf{w}\cdot \mathbf{x}_{i})+b-y_{i} \leq \varepsilon+\xi_{i}^{*} \hspace{-0.8em}\\ 
\ \xi_{i},\xi_{i}^{*} \geq 0, \quad i=1 \ldots n \hspace{-0.8em}\\
\\
\quad \quad \quad (\text{\textcolor{blue}{\textbf{convex}}})
\end{array}
}
\end{split}
+
\boxed{
\begin{array}{lll}
C^{*}\sum\limits_{j=1}^m \zeta_{j}  \hspace{-0.8em}\\ 
\\
\vert y_{j}^{*}-(\mathbf{w}\cdot \mathbf{x}_{j}^{*})-b \vert \geq \Delta-\zeta_{j} \hspace{-0.8em}\\ 
\\
\zeta_{j} \geq 0, \quad j=1 \ldots m \hspace{-0.8em} \\ \\
(\text{\textcolor{red}{\textbf{concave}}}\  U_{\Delta}(y_j-f(\mathbf{x}_{j}^{*})))
\end{array}
}\nonumber
\end{equation}
Here, parameters : $\varepsilon,\Delta \geq 0$ and $C,C^{*} \geq 0$ control the tradeoff between `explanation' of training samples and `falsification' of the universum samples. This new optimization formulation for U-SVR has two additional tuning parameters, $C^{*}$ and $\Delta$, relative to standard SV regression setting. Note that setting $C^{*} =0$ or $\Delta = 0$ in the U-SVR formulation yields standard SVR formulation. \newline

\subsection{Computational Implementation of U-SVR} \label{sec32}

The U-SVR formulation \eqref{eq8} is non-convex due to non-convexity of the Universum loss $U_{\Delta}(y_{j}^{*}-f(\mathbf{x}_{j}^{*}))$ shown in Fig. \ref{fig3}. Hence, it cannot be solved using standard convex solvers commonly used in machine learning.  Recently, similar non-convex optimization problems have been addressed in \cite{collobert06,shen03} using the ConCave Convex Programming (CCCP) strategy. According to CCCP strategy, the cost function $J(\theta)$ is decomposed as the sum of a convex part $J_{vex}(\theta)$ and a concave part $J_{cav}(\theta)$, where  $ \theta$ is the optimization argument. Each iteration of the CCCP procedure approximates the concave part by its tangent and minimizes the resulting convex function (see Algorithm \ref{alg1}).

\begin{algorithm}  
 Initialize $ \theta^{0}$ \;
 \Repeat{convergence of $\theta$}{
 	$\theta^{t+1} = \underset{\theta}{\text{argmin}}\ (J_{vex}(\theta)+J_{cav}^{'}(\theta_t)\cdot \theta)$\;
 }
 \caption{ConCave Convex Programming (CCCP)\label{alg1}}
\end{algorithm} 
Hence, we propose to apply the CCCP strategy for solving the non-convex optimization formulation \eqref{eq8}. Detailed application of the CCCP strategy and the resulting algorithm for solving the U-SVM regression formulation \eqref{eq8} are presented next.

\begin{figure}
\centering
\subfloat[][]{\includegraphics[height=1.6in, width=1.7in]{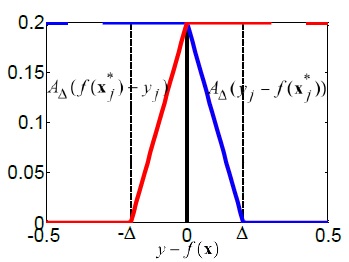}\label{fig4a}}
\subfloat[][]{\includegraphics[height=1.6in, width=1.7in]{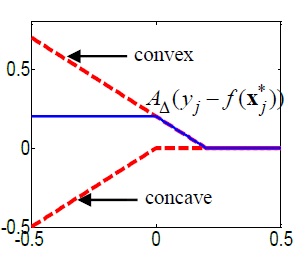}\label{fig4b}}
    \caption{(a)Universum loss as the sum of two ramp losses $A_{\Delta}(y_{j}^{*}-f(\mathbf{x}_{j}^{*}))$ and $A_{\Delta}(f(\mathbf{x}_{j}^{*})-y_{j}^{*})$.(b) Decomposition of $A_{\Delta}(y_{j}^{*}-f(\mathbf{x}_{j}^{*}))$ as the sum of a convex and concave loss.}
\end{figure}

The Universum loss function can be represented as a sum of two ramp losses, $U_{\Delta}(y_{j}^{*}-f(\mathbf{x}_{j}^{*})) =A_{\Delta}(y_{j}^{*}-f(\mathbf{x}_{j}^{*})) + A_{\Delta}(f(\mathbf{x}_{j}^{*})-y_{j}^{*})$+ a constant (see Fig. \ref{fig4a}); where $A_{\Delta}(y_{j}^{*}-f(\mathbf{x}_{j}^{*})) = max(0,\Delta -t)-max(0,-t)$ (see  Fig. 4b). The constant term does not affect the optimization; and hence \eqref{eq8} can be re-written as,
\begin{flalign}\label{eq9}
\underset{\mathbf{w},b,\mathbf{\xi},\mathbf{\xi}^{*},\mathbf{\zeta},\mathbf{\zeta}^{*}}{\text{min}}& \quad \quad \frac{1}{2}(\mathbf{w}\cdot \mathbf{w}) + C\sum\limits_{i=1}^n(\xi_{i} +\xi_{i}^{*})+C^{*}\sum\limits_{j=1}^m(\zeta_{i} +\zeta_{i}^{*}) \nonumber &&\\
&\quad \quad -\sum\limits_{j=1}^{2m} H(y_{j}^{*},f(\mathbf{x}_{j}^{*})) &&\\ 
s.t. & \quad \quad y_{i}-(\mathbf{w}\cdot \mathbf{x_{i}})-b \leq \varepsilon+\xi_{i}, \quad \quad  \xi_{i}\geq 0 \nonumber &&\\ 
& \quad \quad (\mathbf{w}\cdot \mathbf{x}_{i})+b-y_{i} \leq \varepsilon+\xi_{i}^{*}, \quad \quad  \xi_{i}^{*}\geq 0 \nonumber &&\\
& \quad \quad y_{j}^{*}-(\mathbf{w}\cdot \mathbf{x}_{j}^{*})-b \leq -\Delta+\zeta_{j}, \quad \zeta_j \geq 0 \nonumber &&\\
& \quad \quad (\mathbf{w}\cdot \mathbf{x}_{j}^{*})-b -y_{j}^{*} \leq -\Delta+\zeta_{j}^{*}, \quad \zeta_j^{*} \geq 0 \nonumber &&\\
& \quad \quad i=1 \ldots n , \quad j=1 \ldots m && \nonumber
\end{flalign}
where,
\begin{flalign}
H(y_{j}^{*},f(\mathbf{x}_{j}^{*})) &= && \nonumber \\
&\left\{
\begin{array}{l l}      
    max(0,-y_{j}^{*}+(\mathbf{w}\cdot \mathbf{x}_{j}^{*})-b);\  j=1 \ldots m \\
    max(0,y_{j}^{*}-(\mathbf{w}\cdot \mathbf{x}_{j}^{*})-b);\ j=m+1 \ldots 2m
\end{array}\right. && \nonumber 
\end{flalign}
and $f(\mathbf{x})=(\mathbf{w} \cdot \mathbf{x})+b$. Next, define :
\begin{flalign}
k_j &= -C^*\frac{\partial H(y_{j}^{*},f(\mathbf{x}_{j}^{*}))}{\partial f(\mathbf{x}_{j}^{*})}&&\\
&=\left\{
\begin{array}{l l}      
    -C^{*};&  if\quad  y_{j}^{*} <f(\mathbf{x}_{j}^{*});\ j=1 \ldots m \\
    \ C^{*};&  if\quad y_{j}^{*} >f(\mathbf{x}_{j}^{*});\ j=m+1 \ldots 2m \\
    \ 0 ;&\text{else}
\end{array}\right. && \nonumber \\
&\equiv\left\{
\begin{array}{l l}      
    -C^{*};&  if\quad  y_{j}^{*} <f(\mathbf{x}_{j}^{*});\ j=1 \ldots m \\
    \ C^{*};&  if\quad y_{j}^{*} >f(\mathbf{x}_{j}^{*});\ j=1 \ldots m \\
    \ 0 ;&\text{else}
\end{array}\right. && \nonumber
\end{flalign}
The last equivalence follows as the conditions are mutually exclusive.
Hence, \newline \newline
$-C^{*}\frac{\partial H(y_{j}^{*},f(\mathbf{x}_{j}^{*}))}{\partial \theta}\cdot \theta =-C^{*}\frac{\partial H(y_{j}^{*},f(\mathbf{x}_{j}^{*}))}{\partial f(\mathbf{x}_{j}^{*})}\cdot \frac{\partial f(\mathbf{x}_{j}^{*})}{\partial \theta} \cdot \theta $ \newline
$= k_j \left[\begin{array}{l l} \mathbf{x}_{j}^{*}\\ 1 \end{array}\right] ^\top \left[\begin{array}{l l} \mathbf{w}\\ b \end{array}\right]= k_j(\mathbf{w} \cdot \mathbf{x}_{j}^{*} + b)$; \quad with $\theta = \left[\begin{array}{l l} \mathbf{w}\\ b \end{array}\right] $. \newline  \newline Hence, application of the CCCP strategy to solving formulation \eqref{eq8} yields the following Algorithm \ref{alg2}. 

\begin{algorithm} \SetAlgoNoLine 
1. Initialize $ (\mathbf{w}^{0},b^{0})$ using the standard SVR model (see eq. \eqref{eq6}) \;
 \Repeat{convergence i.e. $k_j^{t+1}=k_j^{t} \quad \forall j =1 \ldots m$}{
 	2. At $t+1$ iteration update,\
 	\begin{flalign}
 	k_j^{t+1}&=\left\{
	\begin{array}{l l}      
    -C^{*};\quad  if\quad  y_{j}^{*} <(\mathbf{w}^{t} \cdot \mathbf{x}_{j}^{*})+b^{t};\ j=1 \ldots m \\
    \quad C^{*};\quad  if\quad y_{j}^{*} >(\mathbf{w}^{t} \cdot \mathbf{x}_{j}^{*})+b^{t};\ j=1 \ldots m \\
    \quad 0 ;\quad \text{else}
	\end{array}\right. && \nonumber 
	\end{flalign} \
	3. Solve the following eq. \eqref{eq9} to obtain $(\mathbf{w}^{t+1},b^{t+1})$ \
	\begin{flalign}
\underset{\mathbf{w},b,\mathbf{\xi},\mathbf{\xi}^{*},\mathbf{\zeta},\mathbf{\zeta}^{*}}{\text{min}}& \quad \quad \frac{1}{2}(\mathbf{w}\cdot \mathbf{w}) + C\sum\limits_{i=1}^n(\xi_{i} +\xi_{i}^{*}) \nonumber &&\\
&+C^{*}\sum\limits_{j=1}^m(\zeta_{i} +\zeta_{i}^{*}-k_j^{t+1}(\mathbf{w} \cdot \mathbf{x}_{j}^{*} + b)) \nonumber &&\\ 
&&& \nonumber\\
s.t. & \quad \quad y_{i}-(\mathbf{w}\cdot \mathbf{x_{i}})-b \leq \varepsilon+\xi_{i}, \quad \quad  \xi_{i}\geq 0 \nonumber &&\\ 
& \quad \quad (\mathbf{w}\cdot \mathbf{x}_{i})+b-y_{i} \leq \varepsilon+\xi_{i}^{*}, \quad \quad  \xi_{i}^{*}\geq 0 \nonumber &&\\
& \quad \quad y_{j}^{*}-(\mathbf{w}\cdot \mathbf{x}_{j}^{*})-b \leq -\Delta+\zeta_{j}, \quad \zeta_j \geq 0 \nonumber &&\\
& \quad \quad (\mathbf{w}\cdot \mathbf{x}_{j}^{*})-b -y_{j}^{*} \leq -\Delta+\zeta_{j}^{*}, \quad \zeta_j^{*} \geq 0 \nonumber &&\\
& \quad \quad i=1 \ldots n , \quad j=1 \ldots m && \nonumber
\end{flalign}
 }
 \caption{CCCP algorithm for U-SVR \label{alg2}}
\end{algorithm} 
This Algorithm \ref{alg2} can be extended to nonlinear case by transforming the problem in its dual form (as shown next). \newline Rewrite,
\begin{flalign}
\mathbf{x}_i &= \left\{
\begin{array}{l l l} 
\mathbf{x}_i \quad & i=1 \ldots n \quad \text{(training samples)} \\
\mathbf{x}_j^{*} \quad & i=n+1 \ldots n+m \quad \text{(universum samples)}
\end{array} \right. && \nonumber \\
y_i &= \left\{
\begin{array}{l l l} 
y_i \quad & i=1 \ldots n  \\
y_j^{*} \quad & i=n+1 \ldots n+m
\end{array} \right. && \nonumber \\
\rho_i &= \left\{
\begin{array}{l l l} 
\varepsilon \quad & i=1 \ldots n \\
-\Delta \quad & i=n+1 \ldots n+m
\end{array} \right. &&  \\
C_i &= \left\{
\begin{array}{l l l} 
C \quad & i=1 \ldots n \\
C^{*} \quad & i=n+1 \ldots n+m
\end{array} \right. && \nonumber \\
\delta_i &= \left\{
\begin{array}{l l l} 
C^{*} \quad & if\ y_i<f(\mathbf{x}_i) ;\quad i=n+1 \ldots n+m \\
0 \quad & \text{else}
\end{array} \right. && \nonumber \\
\gamma_i &= \left\{
\begin{array}{l l l} 
C^{*} \quad & if\ y_i>f(\mathbf{x}_i) ;\quad i=n+1 \ldots n+m \\
0 \quad & \text{else}
\end{array} \right. && \nonumber 
\end{flalign}   

Then, we obtain the following Algorithm \ref{alg3} in dual form. The proof follows from standard KKT equations and it is omitted in this paper.
\begin{algorithm}  \SetAlgoNoLine
1. Initialize $ (\mathbf{\alpha}^{0},\mathbf{\beta}^{0},b^{0})$ using the standard SVR model (see eq. \eqref{eq7}) \;
 \Repeat{convergence i.e. $\delta_i^{t+1}=\delta_i^{t}, \quad \gamma_i^{t+1}=\gamma_i^{t} \quad \forall i =n+1 \ldots n+m$}{
 	2. At $t+1$ iteration update,\
 	\begin{flalign}
	\delta_i^{t+1} &= \left\{
	\begin{array}{l l l} 
	C^{*} \quad &  if \quad y_i<\sum \limits_{i=1}^{n+m}(\alpha_i^{t}-\beta_i^{t})+b^{t}  \\
	0 \quad  & \text{else} ;\quad i=n+1 \ldots n+m
	\end{array} \right. && \nonumber \\
	\gamma_i^{t+1} &= \left\{
	\begin{array}{l l l} 
	C^{*} \quad & if \quad y_i>\sum \limits_{i=1}^{n+m}(\alpha_i^{t}-\beta_i^{t})+b^{t}  \\
	0 \quad & \text{else} ;\quad i=n+1 \ldots n+m
	\end{array} \right. && \nonumber 
	\end{flalign}  \
	3. Solve the following eq. \eqref{eq12} to obtain $(\mathbf{\alpha}^{t+1},\mathbf{\beta}^{t+1})$ \
	\begin{flalign} \label{eq12}
	\underset{\mathbf{\alpha},\mathbf{\beta}}{\text{min}} & \quad \quad \frac{1}{2}\sum \limits_{i,j =1}^{n+m} (\alpha_{i} - \beta_{i})(\alpha_{j} - \beta_{j})K(\mathbf{x}_{i} \cdot \mathbf{x}_{j}) &&\\
	& \quad \quad + \sum \limits_{i=1}^{n+m} \rho_i (\alpha_{i} + \beta_{i}) - \sum \limits_{i=1}^{n+m} y_{i}(\alpha_{i} - \beta_{i})&& \nonumber \\
	s.t. & \quad \quad \sum \limits_{i=1}^{n+m} \alpha_{i} = \sum \limits_{i=1}^{n+m} \beta_{i}; \quad i = 1 \ldots n+m \nonumber \\
	 & -\gamma_i^{t+1} \leq \alpha_{i} \leq C_i-\gamma_i^{t+1};\ -\delta_i^{t+1} \leq \beta_{i} \leq C_i-\delta_i^{t+1}    \nonumber
	\end{flalign}
 }
 \caption{CCCP algorithm for U-SVR in dual form \label{alg3}}
\end{algorithm}

This CCCP based non-convex minimization may have local optima, so a good initialization and stopping criteria are critical for this algorithm. In our implementation, standard SVR model is used as the initial condition (as shown in Algorithms \ref{alg2} and \ref{alg3}). Thus the CCCP strategy searches for local minima near the SVR solution. Further, at each iteration we are solving an SVR-like formulation (see eq. \eqref{eq12}). The only difference is in the constraints as shown in eq. \eqref{eq12}. That is, the dual variables in U-SVR formulation \eqref{eq12} have different upper and lower range values, as compared to standard SVR formulation \eqref{eq7}. The time complexity for solving the U-SVR formulation using CCCP is similar to solving standard SVR formulation with $(n+m)$ samples at each iteration. Our preliminary experiments suggests fast convergence (2-5 iterations) for several data sets. Hence, this strategy should be scalable for most real-life datasets.

The kernelized version of U-SVR formulation has five tunable parameters: $C, C^{*}$, kernel parameter, $\varepsilon$ and $\Delta$. So model selection (parameter tuning) becomes an issue for real-life applications. In this paper, we propose simple two-step strategy for U-SVR model selection:
 
\begin{itemize}
\item[1.] \uline{Model selection for standard SVM regression}. This step performs estimation of standard SVM regression model using only training samples. Following \cite{cherkassky13,cherkassky07,cherkassky04}, we can select $C$ parameter analytically, e.g. by setting $C = y_{max} - y_{min}$ , and then perform tuning of $\varepsilon$ and kernel parameters via resampling or separate validation data set. This step performs model selection for tuning parameters specific only to the training samples in the U-SVR formulation in \eqref{eq8}.
\item[2.] \uline{Model selection for tuning two Universum-specific par-\\ameters}. This step performs selection (or tuning) of parameters $C^{*}/C$ and $\Delta$  specific to the U-SVR formulation, while keeping $C,\varepsilon $ and kernel parameters fixed (as obtained in Step 1). This can be performed using a separate validation set or via resampling.
\end{itemize}

\subsection{Generating Universum Data}\label{sec33}
Generally, Universum contains data samples from the same application domain as available training data. For example, for handwritten digit recognition application one can use examples of handwritten letters as Universum data samples, along with examples of handwritten digits used as training data. Note that Universum samples follow a different distribution than the training data. For most real-life applications, Universum data is often available. However, selection of good Universum requires application-domain knowledge and good engineering. Another strategy is to generate synthetic Universum directly from available labeled training data - which is often used for classification setting \cite{cherkassky11,weston06}. Hence, in this section we introduce certain strategies for generating synthetic Universum data under regression setting. Generating such synthetic Universum samples requires minimal domain knowledge and should be applicable to most real life problems (as discussed next).

The notion of generating synthetic universum samples has already been used for binary classification problems \cite{chen09,cherkassky11,sinz08,weston06}. For binary classification, the Universum data belongs to the same $\mathbf{x}$ - space as the training data, but these samples are known not to belong to either of the two classes (`+1' or `-1'). The most popular approach used for generating synthetic universum is called `\textit{random averaging}' or RA, where the Universum samples are generated by randomly selecting one positive and one negative training sample, and then computing their average $\mathbf{x}$- value. Then the  $\mathbf{x}$- values of the generated RA Universum samples would follow a different distribution than either of the two classes.

As discussed in Section \ref{sec31}, Universum samples for regression are \textit{labeled}, unlike unlabeled Universum samples under classification setting. Hence, under regression setting, the distribution of Universum samples can be different from the distribution of $\mathbf{x}$ - values (of the training data) or their $y$ - values or both. All these scenarios contribute to a distribution of $(\mathbf{x},y)$  - values of Universum that is different from the training data distribution. This observation motivates several strategies for generating synthetic universum, as discussed next. \newline
\textbf{Strategy 1}: keep the marginal distributions of $\mathbf{x}$ and $y$ values fixed, but change the underlying conditional distribution $p(y\vert\mathbf{x})$. For example, randomly select any two samples from the training data $(\mathbf{x}_1,y_1)$ and  $(\mathbf{x}_2,y_2)$, such that $y_1 \geq \mu_y $ and $y_2 \leq \mu_y $; where $\mu_y $ is the mean of the $y$ - values of training samples. Next, permute the samples to create two new universum samples, i.e. $(\mathbf{x}_1,y_2)$ and $(\mathbf{x}_2,y_1)$. Following this strategy, the marginal distributions of $\mathbf{x}$, $y$ - values remain the same; but the conditional distribution changes. \newline
\textbf{Strategy 2}: change the distribution of the $y$ - values of training samples. For example, randomly select a training sample $(\mathbf{x},y)$ and then replace its $y$- value as $ y^{\prime} \sim \mathcal{N}(\mu_y,\sigma_y)$ (normal distribution), where $\mu_y$ and $\sigma_y$ are the mean and standard deviation of $y$ - values of the training data. \newline
\textbf{Strategy 3}: change the distribution of the $\mathbf{x}$ - values of the training samples. For example, randomly select a training sample $(\mathbf{x},y)$ and randomly permute the input \textit{features} $\mathbf{x}\rightarrow \mathbf{x}^{\prime}$ to create a new universum sample $(\mathbf{x}^{\prime},y)$. \newline
\textbf{Strategy 4}: change the marginal distributions of both $(\mathbf{x},y)$ - values. For example, randomly select a training sample $(\mathbf{x},y)$ and randomly permute the input \textit{features} $\mathbf{x}\rightarrow \mathbf{x}^{\prime}$ as well as replace the $y$ - value of the sample as, $ y^{\prime} \sim \mathcal{N}(\mu_y,\sigma_y)$; where $\mu_y$ and $\sigma_y$ are the mean and standard deviation of the $y$ - values of the training data. 

Note that each of the above strategies modifies the overall distribution of the $(\mathbf{x},y)$ - values of the training data. So these strategies yield universum data having distribution different from the training data. Hence, such Universum data \textit{should be falsified} by the U-SVR formulation \eqref{eq8}. Next, in Section \ref{sec4}, we show empirical performance comparisons for U-SVR using Universum generated by strategies 1 and 2 only. Empirical results using strategies 3 \& 4 have been omitted due to space constraints.

\section{Empirical Results} \label{sec4}
This section provides empirical results to illustrate the effectiveness of the proposed U-SVR formulation relative to standard SVR. Most examples use synthetic data sets and linear SVM parameterization, in order to clarify the effect of Universum on the prediction performance of SVR. All experiments follow the two-step experimental modeling strategy (presented in Section \ref{sec32}), where optimal standard SVR model is estimated first (using only labeled training data), and then U-SVR model is estimated (using both the training data and Universum data). This experimental strategy simplifies comparisons between standard SVR and U-SVR modeling, and also enables tractable model selection for U-SVR. Further, all empirical comparisons use separate \textit{validation data set} for model selection. That is, the regression model is estimated by fitting the \textit{training data set}, but tuning parameters (for each method) are selected using separate \textit{validation data set}, and prediction performance (of the final regression model) is estimated using independent \textit{test set}. For synthetic data sets, training, validation and test samples are generated according to the same (fixed) distribution, where the input $\mathbf{x}$-values are uniformly distributed in $d$-dimensional input space, and output (response) values represent a target function corrupted by additive noise: $y=t(\mathbf{x})+noise$, where $t(\mathbf{x})$ is a target function and noise is Gaussian. The performance index is the normalized root mean squared error:
\begin{flalign} 
NRMS =& \frac{\sqrt{MSE}}{std(y)} 
\end{flalign} \newline
where, $MSE=\frac{1}{n} \sum_{i=1}^{n} (y- \hat{y})^2$, $n$ = no. of samples, and $\hat{y}$ denotes estimated output values. \newline
Our empirical results show the NRMS error separately for training and test data sets. Of course, only the test error is meaningful for prediction performance comparisons, and the training error is shown mainly for additional understanding of modeling results. Further, following \cite{drucker97}, our empirical results for synthetic data use NRMS error calculated using the true output values $t(\mathbf{x})$(not corrupted by noise), whereas results for real-life data sets use NRMS errors calculated using noisy $y$ -values. 

For all comparisons, prediction performance of U-SVR is compared against two common benchmark methods: standard SVM regression and ridge regression. Hence, these comparisons illustrate possible advantages (or limitations) of introducing Universum into regression modeling.

\subsection{Hypercube dataset} \label{sec41}
Our \textit{first experiment} uses a synthetic 30-dimensional hypercube data set, where each input is uniformly distributed in $[0, 1]$ interval. The output is generated as:  
\begin{align}
y &= x_1+ \ldots + x_5 - x_6 - \ldots -x_{10} + \ldots && \nonumber \\
& \quad +x_{21}+ \ldots + x_{25} - x_{26} - \ldots -x_{30} \quad + \quad \delta && \nonumber
\end{align}
where, the noise is Gaussian: $\delta \sim \mathcal{N}(0,\sigma \mathbf{I}) $. For this data set, we use linear SVR parameterization and consider two different types of universum.\newline
\textit{Universum 1}: input samples follow the same distribution as training samples, e.g., $\mathbf{x} \in \Re^{30}$ and uniformly distributed in $[0,1]$. The output is generated as: 
\begin{align}
y &= -x_1- \ldots - x_5 + x_6 + \ldots +x_{10} + \ldots && \nonumber \\
& \quad -x_{21}- \ldots - x_{25} + x_{26} + \ldots +x_{30} && \nonumber
\end{align}
\textit{Universum 2}: following strategy 2 for generating synthetic universum, we randomly select a training sample $(\mathbf{x},y)$ and re-set its $y$-value as $ y^{\prime} \sim \mathcal{N}(\mu_y,\sigma_y)$, where $\mu_y$ and $\sigma_y$ are the mean and std. deviation of $y$ - values of the training data. \newline
The experimental setting is specified below:
\begin{itemize}
\item[--] No. of training and validation samples = 30, 150 (characterizing low and high sample-size settings, respectively. The number of validation samples is always set to be the same as the number of training samples).
\item[--] No. of test samples = 5000.
\item[--] No. of universum samples = 300.
\item[--] Two additive noise levels  $\sigma$=0.5 and 2 are considered, in order to capture the effects of \textit{low} and \textit{high} noise levels respectively. For the high sample size setting we provide the results for $\sigma$=0.5 only. Experiments involving high noise conditions for high sample size settings did not yield any additional improvement when using the U-SVR formulation.
\end{itemize}
Performance results in Tables \ref{tab1} and \ref{tab2} show the average NRMS error for training and test data observed over 25 random experiments. Here, for each experiment we randomly select the training/validation/test set. The standard deviation of the NRMS values (over 25 experiments) is shown in parenthesis. All comparisons assume linear parameterization for standard SVM regression, U-SVR and ridge regression. These empirical results indicate that for low-sample settings introducing Universum can indeed improve prediction performance, especially for low-noise settings (as shown in Table \ref{tab1}). However, for high-sample size settings introducing Universum does not yield any improvement relative to standard SVR or ridge regression (as evident from Table \ref{tab2}).

\begin{figure}
\centering
\includegraphics[height=1.2in, width=2in]{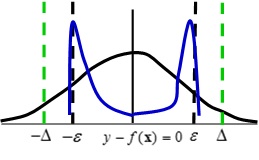}
\caption{Representation of the histogram of residuals $y-f(\mathbf{x})$ for estimated regression model. Training samples are shown in blue and Universum samples shown in black.  The estimated $\pm \varepsilon$ value is shown in black dashed lines and the $\pm \Delta$ value is shown in green dashed line.} \label{fig5}
\end{figure} 

\begin{table}
\centering
\caption{Comparison of average test error for different Universa for low sample size settings $n$ = 30.}  \label{tab1}

\begin{tabular}{|c|c|c|c|c|} \hline
\specialcell{\textbf{NRMS} \\ \textbf{(in \%)}}& \specialcell{\textbf{Ridge}\\\textbf{Regression}}&\textbf{SVR}&\specialcell{\textbf{U-SVR}\\(type 1)}&\specialcell{\textbf{U-SVR}\\(type 2)}\\ \hline
\multicolumn{5}{|c|}{Low noise: no. of training samples $n$=30 with $\sigma$=0.5}\\
\hline
Test & \specialcell{54.56\\(9.32)}& \specialcell{55.03\\(9.11)} & \specialcell{47.79 \\(7.93)} & \specialcell{52.72\\(9.66)}\\
\hline
Training & \specialcell{21.03\\(9.28)}& \specialcell{20.45\\(9.47)} & \specialcell{20.24 \\(8.76)} & \specialcell{20.33\\(8.96)}\\
\hline
\multicolumn{5}{|c|}{High noise: no. of training samples $n$=30 with $\sigma$=2}\\
\hline
Test & \specialcell{97.15\\(10.2)}& \specialcell{97.62\\(11.8)} & \specialcell{92.62 \\(16.3)} & \specialcell{97.41\\(13.9)}\\
\hline
Training & \specialcell{74.12\\(17.83)}& \specialcell{79.36\\(17.19)} & \specialcell{76.55 \\(12.97)} & \specialcell{79.95\\(23.4)}\\
\hline\end{tabular}
\end{table}  

For understanding of the U-SVR modeling results we adopt the technique known as `\textit{histogram of projections}' originally introduced for SVM classification setting \cite{cherkassky10,cherkassky11}. Under classification setting, the `projection value' for a given sample measures its distance from SVM decision boundary. For regression, conceptually similar quantity is the residual $y-f(\mathbf{x})$ that measures the difference between response $y$ and its estimate $f(\mathbf{x})$. So for regression models we use the univariate \textit{histogram of residuals} or residual values $y-f(\mathbf{x})$, where $f(\mathbf{x})$ is the trained regression model with optimally tuned parameters. The typical histogram of residuals of training data for trained SVR model is shown in Fig. \ref{fig5}. In addition, Fig. \ref{fig5} also shows projections of the residual values ($y^{*}-f(\mathbf{x}^{*})$) of universum samples (shown in black). Similar to methodology developed for U-SVM classification, visual interpretation of the histograms of residuals for training data and Universum data (such as shown in Fig. \ref{fig5}) can be used for understanding the effectiveness of Universum under regression setting. In particular, Fig. \ref{fig5} shows the effect of data piling or clustering of residual values for training samples near the boundaries of  $\varepsilon$-tube, which is similar to data piling at the margin borders for SVM classification \cite{cherkassky10,cherkassky11}. This effect of data piling is typically observed for `small-sample' regression data sets corresponding to very ill-posed estimation problems \cite{boyd04}. For such ill-posed settings, introducing additional constraints (in the form of Universum data) usually improves the quality of estimated models. For example, assuming the distribution of residuals for Universum samples (relative to standard SVR model) is as shown in Fig. \ref{fig5}, application of U-SVR is expected to modify/improve the original SVR model by pushing the Universum samples further away from a regression model, according to the Universum loss function in Fig. \ref{fig3}.

\begin{figure}
\centering
\subfloat[][]{\includegraphics[height=1.1in, width=1.6in]{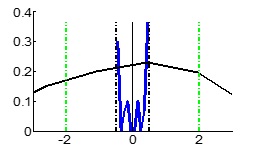}\label{fig6a}}
\subfloat[][]{\includegraphics[height=1.1in, width=1.6in]{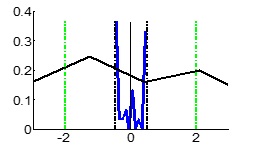}\label{fig6b}}
    \caption{Histogram of residuals for training samples (in blue) and Universum 1 samples (in black) with no. of training samples = 30 and $\sigma = 0.5$. (a)histogram for standard SVR model ($C$= 6.73, $\varepsilon$=0.5) (b) histogram for U-SVR model ($C^{*}/C$ = 0.05, $\Delta$ = 2).}
\end{figure} 

\begin{figure}
\centering
\subfloat[][]{\includegraphics[height=1.1in, width=1.6in]{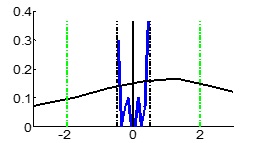}\label{fig7a}}
\subfloat[][]{\includegraphics[height=1.1in, width=1.6in]{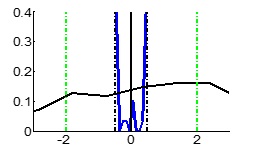}\label{fig7b}}
    \caption{Histogram of residuals for training samples (in blue) and Universum 2 samples (in black) with no. of training samples = 30 and $\sigma = 0.5$. (a)histogram for standard SVR model ($C$= 6.73, $\varepsilon$=0.5) (b) histogram for U-SVR model ($C^{*}/C$ = 0.01, $\Delta$ = 2).}
\end{figure}

Next, we present actual typical histograms for residuals of training samples and Universum samples, for several representative data sets under small-sample size setting, e.g. 30 training samples. Figs. 6 and 7 show the histograms for low noise level, and Figs. 8 and 9 show the histograms for high noise level in the data. For example, Fig. \ref{fig6a} shows the histogram of residuals for optimally trained SVR model (in blue) and the histogram of residuals for Universum (in black). This histogram for training data clearly shows the effect of data piling for training samples, and also shows that the distribution for Universum 1 data is unimodal and centered around the SVR model (marked as the point `0' on x-axis in Fig. \ref{fig6a}). Therefore, we can expect that introducing Universum 1 will change/improve the regression model. This is confirmed by analyzing the histograms of residuals for the trained U-SVR shown in Fig. \ref{fig6b}, which shows the effect of pushing Universum 1 samples away from the estimated regression model (marked as the point `0' on x-axis in Fig. \ref{fig6b}). Specifically, for the SVR model (in Fig.  \ref{fig6a}), the fraction of universum samples lying within the $\pm \Delta$ zone is $\sim$ 92\% and that for the U-SVR model (in Fig. \ref{fig6b}) this fraction is $\sim$ 85\%. Hence, the U-SVR model (in Fig. \ref{fig6b}) increases the contradiction for the universum samples and improves the prediction performance (relative to standard SVR), which is consistent with results in Table \ref{tab1}. Similar reasoning applies to Fig. 7 showing the histograms for the same training data but using type 2 Universum. In this case, however, the fraction of Universum samples lying within the $\pm \Delta$ zone for SVR model (in Fig. \ref{fig7a}) is $\sim$ 90\% and that for the U-SVR model (in Fig. \ref{fig7b}) is $\sim$ 90\%.  So we can expect little or no improvement in prediction performance (relative to SVR), which is confirmed by results in Table \ref{tab1}.

\begin{figure}
\centering
\subfloat[][]{\includegraphics[height=1.1in, width=1.6in]{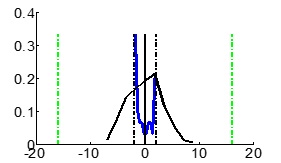}\label{fig8a}}
\subfloat[][]{\includegraphics[height=1.1in, width=1.6in]{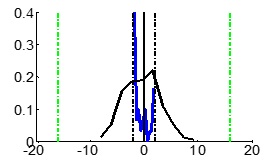}\label{fig8b}}
    \caption{Histogram of residuals for training samples (in blue) and Universum 1 samples (in black) with no. of training samples = 30 and $\sigma = 2$. (a)histogram for standard SVR model ($C$= 14.5, $\varepsilon$=2) (b) histogram for U-SVR model ($C^{*}/C$ = 0.001, $\Delta$ = 16).}
\end{figure} 

\begin{figure}
\centering
\subfloat[][]{\includegraphics[height=1.1in, width=1.6in]{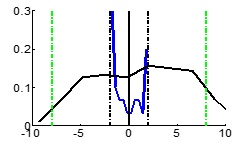}\label{fig9a}}
\subfloat[][]{\includegraphics[height=1.1in, width=1.6in]{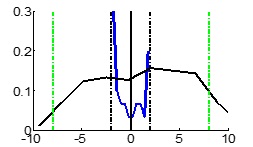}\label{fig9b}}
    \caption{Histogram of residuals for training samples (in blue) and Universum 1 samples (in black) with no. of training samples = 30 and $\sigma = 2$. (a)histogram for standard SVR model ($C$= 14.5, $\varepsilon$=2) (b) histogram for U-SVR model ($C^{*}/C$ = 0.0001, $\Delta$ = 8).}
\end{figure} 

Similarly, we can analyze histograms of residuals for the low sample size, high noise level data shown in Figs. 8 and 9. In this case, the data piling effect for standard SVR model is less strong (as compared with low-noise level data in Figs. 6 and 7). Further, visual comparison of the histograms for Universum data for standard SVR and U-SVR suggests no significant change in the fraction of Universum samples within the $\pm  \Delta$ zone. Hence, we can expect only minor or no improvement in the prediction performance for U-SVR (relative to standard SVR), which is confirmed by results in Table \ref{tab1}.

Finally, consider large sample size, low noise training data set ($n$ = 150, $\sigma$ = 0.5). Having large number of training samples yields very accurate SVR model estimation. For such data sets, we do not observe the data piling effect at the $\pm \varepsilon$  values (see Figs. 10 \& 11), so we expect no improvement from introducing Universum data (as evident from results shown in Table \ref{tab2}).

\begin{figure}
\centering
\subfloat[][]{\includegraphics[height=1.1in, width=1.6in]{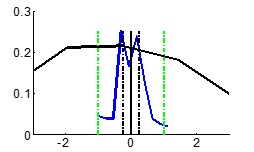}\label{fig10a}}
\subfloat[][]{\includegraphics[height=1.1in, width=1.6in]{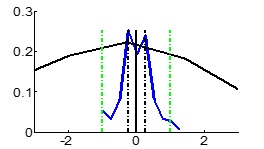}\label{fig10b}}
    \caption{Histogram of residuals for training samples (in blue) and Universum 1 samples (in black) with no. of training samples = 150 and $\sigma = 0.5$. (a)histogram for standard SVR model ($C$= 9.65, $\varepsilon$=0.25) (b) histogram for U-SVR model ($C^{*}/C$ = 0.5, $\Delta$ = 1).}
\end{figure} 

\begin{figure}
\centering
\subfloat[][]{\includegraphics[height=1.1in, width=1.6in]{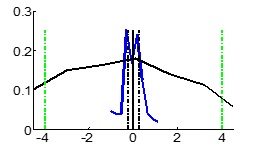}\label{fig11a}}
\subfloat[][]{\includegraphics[height=1.1in, width=1.6in]{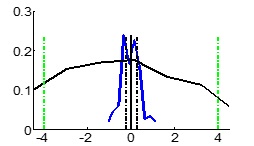}\label{fig11b}}
    \caption{Histogram of residuals for training samples (in blue) and Universum 1 samples (in black) with no. of training samples = 150 and $\sigma = 0.5$. (a)histogram for standard SVR model ($C$= 9.65, $\varepsilon$=0.25) (b) histogram for U-SVR model ($C^{*}/C$ = 0.1, $\Delta$ = 4).}
\end{figure} 

As evident from experiments above, U-SVR is particularly effective for very sparse settings ($\sim$ high-dimensional and low sample size) under low noise conditions. Under such settings, the training data exhibits large data-piling effect near the $\pm \varepsilon$ margins for the SVR model, and introducing the Universum usually helps to improve the prediction performance. With increased noise level (in the data), the data-piling effect becomes less prominent and the U-SVR yields no notable improvement over the SVR solution. Finally, when the number of training samples is large, the estimation problem becomes well-posed and standard SVR model does not exhibit any data-piling effect at the   margins. In this case, application of U-SVR does not provide any improvement over standard SVR. 

The next experiment follows the same experimental set up for 30-dimensional hypercube data set under low-sample size setting as above, except that the additive noise level is set to zero ($\sigma$= 0). Based on our previous discussion, this data set is expected to show the most significant improvement in prediction performance of U-SVR (relative to standard SVR).  Table \ref{tab3} shows performance comparisons, suggesting that U-SVR (using Universum 1) indeed provides very significant improvement over standard SVR solution. 
\begin{table} 
\centering
\caption{Comparison of average test error for different Universa for high sample size settings $n$ =150.} \label{tab2}
\begin{tabular}{|c|c|c|c|c|} \hline
\specialcell{\textbf{NRMS} \\ \textbf{(in \%)}}& \specialcell{\textbf{Ridge}\\\textbf{Regression}}&\textbf{SVR}&\specialcell{\textbf{U-SVR}\\(type 1)}&\specialcell{\textbf{U-SVR}\\(type 2)}\\ \hline
\multicolumn{5}{|c|}{Low noise: no. of training samples $n$=150 with $\sigma$=0.5}\\
\hline
Test & \specialcell{15.83\\(0.85)}& \specialcell{16.19\\(2.02)} & \specialcell{16.43 \\(2.12)} & \specialcell{15.98\\(2.18)}\\
\hline
Training & \specialcell{14.72 \\(1.61)}& \specialcell{14.61\\(1.26)} & \specialcell{14.92 \\(1.28)} & \specialcell{14.41\\(1.41)}\\
\hline\end{tabular}
\end{table} 

\begin{table} 
\centering
\caption{Comparison of average test error for different Universa for $n$ =30 with zero noise.} \label{tab3}

\begin{tabular}{|c|c|c|c|c|} \hline 
\specialcell{\textbf{NRMS} \\ \textbf{(in \%)}}& \specialcell{\textbf{Ridge}\\\textbf{Regression}}&\textbf{SVR}&\specialcell{\textbf{U-SVR}\\(type 1)}&\specialcell{\textbf{U-SVR}\\(type 2)}\\ \hline
\multicolumn{5}{|c|}{Low noise: no. of training samples $n$=30 with $\sigma$=0.5}\\
\hline
Test & \specialcell{15.99\\(10.33)}& \specialcell{17.51\\(10.23)} & \specialcell{7.29 \\(6.95)} & \specialcell{17.41\\(10.16)}\\
\hline
Training & \specialcell{0.15 \\(0.01)}& \specialcell{1.6\\(2.4)} & \specialcell{0.4 \\(1.0)} & \specialcell{1.2\\(1.6)}\\
\hline\end{tabular}
\end{table}

Finally, the last set of experiments demonstrates how the generalization performance of U-SVR is affected by the number of Universum data samples. Let us consider the same synthetic hypercube data set under small sample size, low noise setting, where the size of Universum data set varies as $m$ = 50, 100, 300, 500. Table \ref{tab4} shows performance comparisons between SVR vs. U-SVR, suggesting that:
\begin{itemize}
\item[--] \textit{for Universum 1}:  prediction performance of U-SVR improves with the number universum samples. However, increasing the number of universum samples above certain value ($\sim$300) does not provide additional improvement.
\item[--] \textit{for Univerum 2}: increasing the number of universum samples does not provide any improvement.
\end{itemize}

These results indicate that for sufficiently large number of universum samples the effectiveness of U-SVR depends mostly on the \textit{type} ($\sim$statistical characteristics) of the universa. Similar to classification settings \cite{cherkassky11,sinz08}, the effectiveness of Universum for regression problems depends on the statistical characteristics of both the training data and the Universum data. Hence, there is a need for additional research on the characterization of `\textit{good}' universum data sets (similar to practical conditions for classification in \cite{chen09,cherkassky11,sinz08}). Such practical conditions for the effectiveness of U-SVR are open for future research.

\subsection{Computer Hardware Dataset} \label{sec42}
Our next experiment uses the publicly available real-life \textit{Computer Hardware dataset} \cite{lichman13}. The goal is to predict the \textit{published relative CPU performance} using several other CPU properties. As preprocessing, the categorical variable \textit{vendor\_name} has been transformed to a binary representation. Further the $y$- values have been scaled as $log(1+y)$ (see \cite{lichman13}). In this experiment, we use two types of synthetic universum generated directly from the training data, following Strategy 1 (Universum 1) and Strategy 2 (Universum 2) as described in Section \ref{sec33}. \newline
The experimental set up is summarized next:
\begin{itemize}
\item[--] No. of training samples = 50.
\item[--] No. of validation samples = 50. (This independent validation samples is used for model selection).
\item[--] No. of test samples =109.
\item[--] No. of universum samples = 100. (Increasing the number of universum samples does not provide additional improvement)
\item[--] No. of input variables = 36.
\end{itemize}

As a part of pre-processing, the $\mathbf{x}$- values of the training data have been pre-scaled uniformly to the same range $[-1, 1]$. Such a pre-scaling of inputs (to the same range) is typically required for SVM modeling. Model selection for SVR and U-SVR is performed over the range of parameters, $C=y_{max} - y_{min}, \varepsilon = [0,2^{-1},\ldots, 2^3], C^{*}/C = [2^{-4},\ldots,2^4]$ and $\Delta = [2^{-4},\ldots,2^4]$ and follows the two-step strategy presented in Section \ref{sec3}. Prediction performance results are calculated for 25 random partitionings of the data into training, validation and test data sets.  Table \ref{tab5} provides the average \textit{NRMS} and \textit{MSE} errors for training and test data sets (averaged over 25 random partitioning of the data) and the standard deviation (shown in parenthesis). For most experiments, the typical optimal SVR model parameters are $C \sim 4.5,\varepsilon = 1$. Likewise, typical optimal parameters for U-SVR when using Universum 1 are $C^{*}/C = 2^{-2} - 2^0, \Delta = 0.25 $; and when using Universum 2 are $C^{*}/C = 2^{-6}, \Delta = 0.5 $. For this dataset, preliminary experiments showed significant data-piling effects for the SVR model. Hence, we can expect improved generalization for the U-SVR model. As evident from Table \ref{tab5}, U-SVR (using Universum 1) provides improved generalization over the standard SVR, whereas Universum 2 yields only minor improvement. For this data set, the ridge regression provided performance similar to standard SVR and hence it has not been reported.

\begin{table} 
\centering
\caption{Comparison of average test error for different Universa with increase in Universum samples.} \label{tab4}
\begin{tabular}{|c|c|c|c|c|} \hline 
\textbf{NRMS}&\multicolumn{4}{|c|}{\textbf{Number of Universum samples}}\\ \cline{2-5}
(in \%) & \quad m=50 \quad & \quad m=100 \quad & \quad m=300 \quad & \quad m=500 \quad \\ \cline{2-4}
\hline
\textbf{SVR}& \specialcell{56.98\\(9.06)}& - & - & - \\
\hline
\specialcell{\textbf{U-SVR}\\(type1)}& \specialcell{55.31\\(8.55)}& \specialcell{52.7\\(6.56)} & \specialcell{47.05\\(9.01)} & \specialcell{47.03\\(9.1)} \\
\hline
\specialcell{\textbf{U-SVR}\\(type2)}& \specialcell{55.03\\(11.07)}& \specialcell{56.75\\(9.8)} & \specialcell{56.58\\(9.09)} & \specialcell{56.32\\(9.12)} \\
\hline
\end{tabular}
\end{table}

\begin{table} 
\centering
\caption{Comparison of SVR vs. U-SVR for CPU data.} \label{tab5}

\begin{tabular}{|c|c|c|c|} \hline 
\multicolumn{4}{|c|}{\textbf{Test Data}}\\
\hline
\specialcell{\quad \quad \textbf{Methods}} \quad \quad & \quad \textbf{SVR} \quad &\quad \specialcell{\textbf{U-SVR}\\(type 1)} \quad &\quad \specialcell{\textbf{U-SVR}\\(type 2)} \quad \\ \hline
\textbf{NRMS(\%)} & \specialcell{58.8\\(13.0)} & \specialcell{53.35\\(7.18)} & \specialcell{56.85 \\(9.8)} \\
\hline
\textbf{MSE} & \specialcell{0.39 \\(0.23)}& \specialcell{0.30\\(0.09)} & \specialcell{0.35 \\(0.15)} \\
\hline
\multicolumn{4}{|c|}{\textbf{Training Data}}\\
\hline
\textbf{NRMS(\%)} & \specialcell{45.7\\(12.3)} & \specialcell{41.8\\(7.8)} & \specialcell{45.1 \\(11.6)} \\
\hline
\textbf{MSE} & \specialcell{0.22 \\(0.1)}& \specialcell{0.18 \\(0.06)} & \specialcell{0.21 \\(0.09)} \\
\hline \end{tabular}
\end{table}

\subsection{Vilmann's Rat Dataset} \label{sec43}
This data set illustrates the effectiveness of U-SVR under non-linear SVR parameterization. The real-life \textit{Vilmann's Rat} dataset contains the skull X-ray images of 21 different rats, represented using 8 - landmarks for 2-dimensions \cite{bookstein97}. The skull X-ray images are available for each rat at ages 7, 14, 21, 30, 40, 60, 90 and 150 days.   The task is to predict the ontogenetic development (\textit{age}) of a rat using the X-ray images. The dataset contains 4 missing data which has been removed from the analysis. The processed dataset contains 164 samples. In this experiment, we use two types of synthetic universum generated directly from the training data, following Strategy 1(Universum 1) and Strategy 2(Universum 2) as described in Section \ref{sec33}. The experimental set up is summarized next:
\begin{itemize}
\item[--] No. of training samples $\sim$ skull x-ray images of 5 different rats. (= 40 images)
\item[--] No. of validation samples $\sim$ skull x-ray images of 5 different rats (= 40 images. This independent validation data set is used for model selection).
\item[--] No. of test samples $\sim$ skull x-ray images of remaining 11 rats. (= 88 images)
\item[--] No. of universum samples = 200 (increasing the number universum samples does not provide additional improvement).
\item[--] Dimensionality of each sample = 16. ($\sim$8 landmarks per 2D)
\end{itemize}

\begin{table} 
\centering
\caption{Comparison of SVR vs. U-SVR for Rat dataset.} \label{tab6}
\begin{tabular}{|c|c|c|c|} \hline 
\multicolumn{4}{|c|}{\textbf{Test Data}}\\
\hline
\specialcell{\quad \quad \textbf{Methods}} \quad \quad & \quad \textbf{SVR} \quad &\quad \specialcell{\textbf{U-SVR}\\(type 1)} \quad &\quad \specialcell{\textbf{U-SVR}\\(type 2)} \quad \\ \hline
\textbf{NRMS(\%)} & \specialcell{26.04\\(3.03)} & \specialcell{25.67\\(2.64)} & \specialcell{ 24.88 \\(2.77)} \\
\hline
\textbf{MSE} & \specialcell{135.3 \\(32.7)}& \specialcell{131.0\\(27.6)} & \specialcell{123.1 \\(27.3)} \\
\hline
\multicolumn{4}{|c|}{\textbf{Training Data}}\\
\hline
\textbf{NRMS(\%)} & \specialcell{17.25\\(4.05)} & \specialcell{17.6\\(3.87)} & \specialcell{17.42 \\(4.03)} \\
\hline
\textbf{MSE} & \specialcell{61.9 \\(25.5)}& \specialcell{64.02 \\(26.9)} & \specialcell{63.08 \\(27.03)} \\
\hline \end{tabular}
\end{table}

For this data set, using RBF kernel of the form $K(\mathbf{x}_i,\mathbf{x}_j) = exp(-\gamma \Vert \mathbf{x}_i - \mathbf{x}_j \Vert^2)$ provided optimal results for SVR modeling. The model selection is done over the range of parameters, $C=y_{max} - y_{min}$, $\gamma = [2^{-6},\ldots, 2^0]$, $\varepsilon = [0, 2^{-4},\ldots,2^6]$, $C^{*}/C = [0,2^{-7},\ldots,2^1] $ and $\Delta = [2^{-4},\ldots,2^4]$ . Typical optimal parameters for SVR selected through model selection are  $C = 143$, $\gamma = 2^{-3} - 2^{-2}$, $\varepsilon = 0, 4, 8 $ (observed high variance). Optimal tuning parameters specific to U-SVR when using Universum 1 are: $ C^{*}/C \sim 2^{-3}-2^{-1}$, $\Delta = 2^5 - 2^6 $; and when using Universum 2 are: $C^{*}/C \sim 0.5$, $\Delta = 2^5 - 2^6$. Table \ref{tab6} provides the average \textit{NRMS} and \textit{MSE} for training and test data sets (calculated over 25 random partitionings of the data). The standard deviation is provided in parenthesis. Preliminary experiments showed data-piling effects for the SVR model. Hence, we can expect improved generalization for the U-SVR model.  This is confirmed by results presented in Table \ref{tab6}, where U-SVR (using Universum 2) provides improved prediction.  

\section{Conclusions} \label{sec5}
This paper extends the idea of universum learning to regression problems and provides new optimization formulation called Universum Support Vector Regression (U-SVR). This U-SVR formulation is non-convex and cannot be solved using standard convex solvers typically adopted for existing SVM software packages. This paper adopts the method of Con-Cave Convex Problem (CCCP) and provides a new algorithm (Algorithm 1 \& 2) for solving the proposed U-SVR formulation. Following this strategy, the current U-SVR formulation can be solved by several iterations of standard SVM-like optimization problem.

Moreover, the proposed U-SVR formulation has 5 tunable parameters: $C$, $C^{*}$, $\varepsilon$, $\Delta$ and kernel parameter. Hence a successful practical application of U-SVR depends on the optimal selection of these model parameters. We propose simple two-step strategy for model selection where optimal model parameters for standard SVR are estimated first, and then model selection for U-SVR involves tuning only two remaining parameters $C^{*}/C$ and $\Delta$. Such a two-step strategy significantly simplifies model selection for U-SVR.

Finally, the paper provides empirical results to show the effectiveness of the proposed U-SVR over standard SVR. Additional results showing the effectiveness of U-SVR are available for several real-life datasets, and have been omitted for space constraints. Our results suggest that U-SVR is particularly effective for high-dimension low (training) sample size settings. Under such settings, the SVR model exhibits significant data piling of the training samples near the $\pm \varepsilon$ margin. For such ill-posed settings, introducing the Universum can provide improved generalization over the standard SVR solution. However, the effectiveness of U-SVR depends both on the statistical characteristics of both the training data and Universum data. These statistical characteristics can be conveniently captured using the `histogram-of-residuals' method introduced in this paper.           

\bibliographystyle{abbrv}
\bibliography{uSVR}

\begin{thebibliography}{10}

\bibitem{bookstein97}
F.~L. Bookstein.
\newblock {\em Morphometric tools for landmark data: geometry and biology}.
\newblock Cambridge University Press, 1997.

\bibitem{boyd04}
S.~Boyd and L.~Vandenberghe.
\newblock {\em Convex Optimization}.
\newblock Cambridge University Press, New York, NY, USA, 2004.

\bibitem{chen09}
S.~Chen and C.~Zhang.
\newblock Selecting informative universum sample for semi-supervised learning.
\newblock In {\em IJCAI}, pages 1016--1021, 2009.

\bibitem{cherkassky13}
V.~Cherkassky.
\newblock {\em Predictive Learning}.
\newblock VCtextbook, 2013.

\bibitem{cherkassky10}
V.~Cherkassky and S.~Dhar.
\newblock Simple method for interpretation of high-dimensional nonlinear svm
  classification models.
\newblock In R.~Stahlbock, S.~F. Crone, M.~Abou-Nasr, H.~R. Arabnia,
  N.~Kourentzes, P.~Lenca, W.-M. Lippe, and G.~M. Weiss, editors, {\em DMIN},
  pages 267--272. CSREA Press, 2010.

\bibitem{cherkassky11}
V.~Cherkassky, S.~Dhar, and W.~Dai.
\newblock Practical conditions for effectiveness of the universum learning.
\newblock {\em Neural Networks, IEEE Transactions on}, 22(8):1241--1255, 2011.

\bibitem{cherkassky04}
V.~Cherkassky and Y.~Ma.
\newblock Practical selection of svm parameters and noise estimation for svm
  regression.
\newblock {\em Neural networks}, 17(1):113--126, 2004.

\bibitem{cherkassky07}
V.~Cherkassky and F.~M. Mulier.
\newblock {\em Learning from Data: Concepts, Theory, and Methods}.
\newblock Wiley-IEEE Press, 2007.

\bibitem{collobert06}
R.~Collobert, F.~Sinz, J.~Weston, and L.~Bottou.
\newblock Large scale transductive svms.
\newblock {\em The Journal of Machine Learning Research}, 7:1687--1712, 2006.

\bibitem{dhar15}
S.~Dhar and V.~Cherkassky.
\newblock Development and evaluation of cost-sensitive universum-svm.
\newblock {\em Cybernetics, IEEE Transactions on}, 45(4):806--818, 2015.

\bibitem{drucker97}
H.~Drucker, Chris, B.~L. Kaufman, A.~Smola, and V.~Vapnik.
\newblock Support vector regression machines.
\newblock In {\em Advances in Neural Information Processing Systems 9},
  volume~9, pages 155--161, 1997.

\bibitem{lichman13}
M.~Lichman.
\newblock Uci machine learning repository.
\newblock \url{http://archive.ics.uci.edu/ml}.
\newblock Accessed: 2016-02-05.

\bibitem{lu14}
S.~Lu and L.~Tong.
\newblock Weighted twin support vector machine with universum.
\newblock {\em Advances in Computer Science: an International Journal},
  3(2):17--23, 2014.

\bibitem{qi14}
Z.~Qi, Y.~Tian, and Y.~Shi.
\newblock A nonparallel support vector machine for a classification problem
  with universum learning.
\newblock {\em Journal of Computational and Applied Mathematics}, 263:288--298,
  2014.

\bibitem{shen12}
C.~Shen, P.~Wang, F.~Shen, and H.~Wang.
\newblock $\{$cal U$\}$ boost: Boosting with the universum.
\newblock {\em Pattern Analysis and Machine Intelligence, IEEE Transactions
  on}, 34(4):825--832, 2012.

\bibitem{shen03}
X.~Shen, G.~C. Tseng, X.~Zhang, and W.~H. Wong.
\newblock {On psi-Learning}.
\newblock {\em Journal of the American Statistical Association}, 98:724--734,
  Jan. 2003.

\bibitem{sinz08}
F.~Sinz, O.~Chapelle, A.~Agarwal, and B.~Sch{\"o}lkopf.
\newblock An analysis of inference with the universum.
\newblock In {\em Advances in neural information processing systems 20}, pages
  1369--1376, Red Hook, NY, USA, Sept. 2008. Max-Planck-Gesellschaft, Curran.

\bibitem{vapnik06}
V.~Vapnik.
\newblock {\em {Estimation of Dependences Based on Empirical Data (Information
  Science and Statistics)}}.
\newblock Springer, Mar. 2006.

\bibitem{vapnik98}
V.~N. Vapnik.
\newblock {\em Statistical Learning Theory}.
\newblock Wiley-Interscience, 1998.

\bibitem{wang14}
Z.~Wang, Y.~Zhu, W.~Liu, Z.~Chen, and D.~Gao.
\newblock Multi-view learning with universum.
\newblock {\em Knowledge-Based Systems}, 70:376--391, 2014.

\bibitem{weston06}
J.~Weston, R.~Collobert, F.~Sinz, L.~Bottou, and V.~Vapnik.
\newblock Inference with the universum.
\newblock In {\em Proceedings of the 23rd international conference on Machine
  learning}, pages 1009--1016. ACM, 2006.

\bibitem{zhang08}
D.~Zhang, J.~Wang, F.~W. 0001, and C.~Zhang.
\newblock Semi-supervised classification with universum.
\newblock In {\em SDM}, pages 323--333. SIAM, 2008.

\end{thebibliography}

\end{document}